\documentclass[10pt,conference]{IEEEtran}
\IEEEoverridecommandlockouts

\usepackage[a4paper, hmargin=13mm, vmargin={25mm,30mm}, headsep=4mm]{geometry}

\usepackage{cite}
\usepackage{amsmath,amssymb,amsfonts}
\usepackage{algorithmic}
\usepackage{graphicx}
\usepackage{textcomp}
\usepackage{xcolor}
\usepackage{svg}
\usepackage{url}
\usepackage{empheq}
\usepackage[utf8]{inputenc}
\DeclareUnicodeCharacter{2010}{-}
\usepackage[colorinlistoftodos]{todonotes}
\def\BibTeX{{\rm B\kern-.05em{\sc i\kern-.025em b}\kern-.08em
    T\kern-.1667em\lower.7ex\hbox{E}\kern-.125emX}}

\usepackage{scalerel}
\usepackage{tikz}
\usetikzlibrary{svg.path}
\definecolor{orcidlogocol}{HTML}{A6CE39}
\tikzset{
  orcidlogo/.pic={
    \fill[orcidlogocol] svg{M256,128c0,70.7-57.3,128-128,128C57.3,256,0,198.7,0,128C0,57.3,57.3,0,128,0C198.7,0,256,57.3,256,128z};
    \fill[white] svg{M86.3,186.2H70.9V79.1h15.4v48.4V186.2z}
                 svg{M108.9,79.1h41.6c39.6,0,57,28.3,57,53.6c0,27.5-21.5,53.6-56.8,53.6h-41.8V79.1z M124.3,172.4h24.5c34.9,0,42.9-26.5,42.9-39.7c0-21.5-13.7-39.7-43.7-39.7h-23.7V172.4z}
                 svg{M88.7,56.8c0,5.5-4.5,10.1-10.1,10.1c-5.6,0-10.1-4.6-10.1-10.1c0-5.6,4.5-10.1,10.1-10.1C84.2,46.7,88.7,51.3,88.7,56.8z};
  }
}
\newcommand\orcidicon[1]{\href{https://orcid.org/#1}{\mbox{\scalerel*{
\begin{tikzpicture}[yscale=-1,transform shape]
\pic{orcidlogo};
\end{tikzpicture}
}{|}}}}
\usepackage{hyperref} 

\newcommand{\rrr}{{\textit{$R^3$}} }

\newcommand\copyrighttext{%
  \footnotesize \textcopyright 2021 IEEE.  Personal use of this material is permitted.  Permission from IEEE must be obtained for all other uses, in any current or future media, including reprinting/republishing this material for advertising or promotional purposes, creating new collective works, for resale or redistribution to servers or lists, or reuse of any copyrighted component of this work in other works.}
\newcommand\copyrightnotice{%
\begin{tikzpicture}[remember picture,overlay]
\node[anchor=south,yshift=10pt] at (current page.south) {\fbox{\parbox{\dimexpr\textwidth-\fboxsep-\fboxrule\relax}{\copyrighttext}}};
\end{tikzpicture}%
}

\begin{document}

\title{
Risk Ranked Recall: Collision Safety Metric for Object Detection Systems in Autonomous Vehicles
}

\author{
  \IEEEauthorblockN{
    Ayoosh Bansal\IEEEauthorrefmark{1}\orcidicon{0000-0002-4848-6850},
    Jayati Singh\IEEEauthorrefmark{1}\orcidicon{0000-0003-1528-7369},
    Micaela Verucchi\IEEEauthorrefmark{2}\IEEEauthorrefmark{3}\orcidicon{0000-0003-3898-8571},
    Marco Caccamo\IEEEauthorrefmark{2},
    Lui Sha\IEEEauthorrefmark{1}
  }
  
  \IEEEauthorblockA{
    \IEEEauthorrefmark{1}University of Illinois Urbana-Champaign,
    \{ayooshb2, jayati, lrs\}@illinois.edu
  }
  \IEEEauthorblockA{
  \IEEEauthorrefmark{2}Technical University of Munich,
 \{micaela.verucchi, mcaccamo\}@tum.de
  }
    \IEEEauthorblockA{
  \IEEEauthorrefmark{3}University of Modena and Reggio Emilia
 }
}

\maketitle
\IEEEpeerreviewmaketitle
\copyrightnotice

\begin{abstract}

Commonly used metrics for evaluation of object detection systems (precision, recall, mAP) do not give complete information about their suitability of use in safety critical tasks, like obstacle detection for collision avoidance in Autonomous Vehicles (AV).
This work introduces the Risk Ranked Recall ($R^3$) metrics for object detection systems.
The \rrr metrics categorize objects within three ranks. Ranks are assigned based on an objective cyber-physical model for the risk of collision. Recall is measured for each rank.

\end{abstract}


\section{Introduction}
\label{sec:introduction}

Obstacle detection is a safety critical perception requirement for AV.
In recent times, there have been great improvements in deep learning based solutions for object detection.
A result and catalyst of this phenomenal progress have been the real world datasets~\cite{bdd100k,kitti_1,waymo_open_dataset} used for development and evaluation of object detection systems.
While plenty of AV datasets have been released, the metrics used with these datasets have not been tailored to the autonomous driving application.
The commonly used metrics for evaluating object detection systems are precision and recall~\cite{recall_precision}.
While these metrics work well for general applications, there remains a need for safety aware object detection metrics~\cite{kpi_needed,ohn2017all}.
To understand this need, let's first consider the two basic metrics, Precision and Recall, used to evaluate the efficacy of an object detection system and apply them to a driving scenario.

{\small
\begin{align}
    \text{Precision} &= \frac{\text{True Positives}}{(\text{True Positives} + \text{False Positives})} \label{eq:pr} \\
    \text{Recall} &= \frac{\text{True Positives}}{(\text{True Positives} + \text{False Negatives})}
    \label{eq:recall}
\end{align}
}

\begin{figure}[tb]
  \centering
  \includegraphics[width=.65\linewidth]{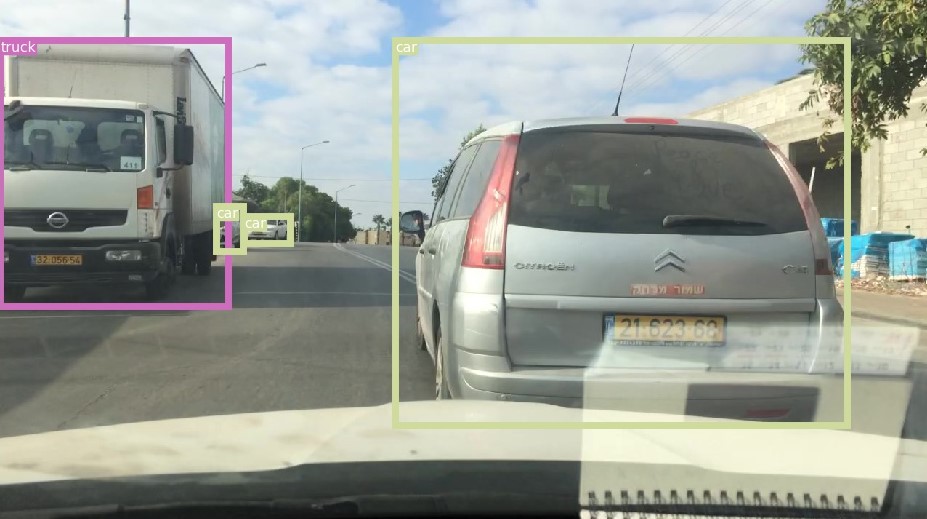}
  \caption{\label{fig:bdd_dets_motivation}
    A front view scene from BDD100K\cite{bdd100k} dataset with 4 labeled vehicles.
    Intuitively, the closer vehicles are more important to detect than those farther away.
    However existing metrics do not consider such a notion of importance.
    \rrr metrics fill that gap and provide risk or importance aware information about behavior of object detection systems.
    }
\end{figure}

Figure~\ref{fig:bdd_dets_motivation} shows a front view image from the ego vehicle\footnote{Ego vehicle is the commonly used term for the first person vehicle.}, annotated with bounding boxes around all other visible vehicles.
If an object detection system correctly detects exactly 2 ($TP = 2$) out of the 4 ($FN = 2$) objects in this image the Recall (Eq.~\ref{eq:recall}) for this result is 0.5.
This Recall value is independent of which 2 vehicles of the 4 were correctly detected.
Intuitively, it is evident that at this time instance, being able to detect the car right in front and the truck coming down the opposite direction is far more important than the two cars further down.
The ego vehicle, controlled by a human or autonomous agent, could end up in an unsafe situation quickly if the nearer car or truck are not detected. At this moment being unable to detect the two cars further away poses little to no risk.

This work introduces the \rrr metrics, 
a risk aware version of Recall.
The risk rankings are based on the risk of collision~($\S$\ref{sec:rrr}).
Recall is measured for each rank separately.

\begin{itemize}
    \item \textit{$R^3_1$} : Recall for objects that pose an \textit{imminent} risk of collision~($\S$\ref{sec:imminent_collision}).
    \item \textit{$R^3_2$} : Recall for objects that can \textit{potentially} collide with the ego vehicle
    ($\S$\ref{sec:potential_collision}).
    \item \textit{$R^3_3$} : All other objects in the environment.
\end{itemize}
By measuring the \rrr values separately, the metrics decouple and emphasise fulfillment of safety requirements.
This work describes the safety aware \textit{Risk Ranked Recall} metrics.
Tools and examples from this work will be made available.\footnote{https://gitlab.engr.illinois.edu/rtesl/risk-ranked-recall}

\section{related work}
\label{sec:relatedwork}

To the best of our knowledge there are no existing safety aware metrics for object detection based on an objective model.
Datasets like KITTI~\cite{kitti_1} and Waymo Open Dataset~\cite{waymo_open_dataset} employ a notion of difficulty.
However, difficulty is based on how difficult it is to detect objects, based on their size, truncation, occlusion or judgement of the groundtruth annotator. There is no consideration for the importance of detecting that object for continued safe operation of the ego vehicle.

Ohn-Bar and Trivedi's object importance for driving is the closest related work~\cite{ohn2017all}.
They pioneered the notion of object importance in object detection.
They asked human drivers to annotate the importance of objects in driving scenes and used the annotations to showcase the advantages of importance guided Deep Neural Network (DNN) training.
There are key differences between their work and ours, primarily due to the differences in intended use.
Their notion of importance is based upon what human drivers consider important while we focus on what obstacles can physically pose a risk of collision.
For safety an approximate notion of importance is not reliable, thus we use an objective model ($\S$\ref{sec:rrr}) for risk.

Neural networks to calculate the threat or risk of different scenes have also been explored~\cite{wang2017collision,feth2018dynamic}.
The main feature is that these DNN can estimate the risk of a situation based on visual input.
They could hence be used for online threat assessment to aid path planning.
However, the approximate nature of these methods makes them unsuitable for our use case.
They are also currently limited to predict the risk posed by a scene and not individual objects.

An alternate approach exists to the objective model proposed in this work for assigning risk ranks to objects.
Objects can be prioritized by measuring the impact of an object on planning decisions~\cite{refaat2019agent}.
While the technique works well for real time agent prioritization, we argue that the model proposed in this work is superior for safety critical evaluations.
First, it does not include any potential bias from specific planning algorithms.
And second, our model considers only safety requirements \textit{i.e.} collision avoidance while the planner based simulations can implicitly conflate mission requirements with safety requirements.

\section{Design Choices}
\label{sec:sysmodel}

Before presenting the risk model, we first lay out the design choices and corresponding reasoning.

\textbf{Time Horizon:}
The analysis need to be time limited because given infinite time any two independent objects can have the potential to collide.
Emergency stop is considered an acceptable emergency response for AV~\cite{meder2019should}.
Hence, a time horizon equal to Time To Stop (TTS) of the ego vehicle is used to limit the risk analysis~(Eq. \ref{eq:tts}).

\begin{figure}[tp]
\centering
\includegraphics[width=.75\linewidth,keepaspectratio]{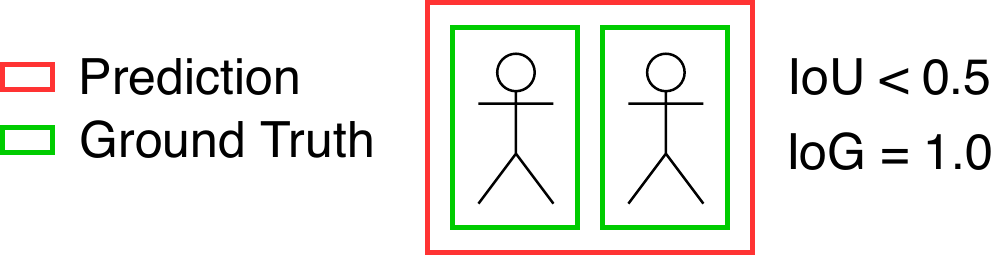}
\caption{\label{fig:uvsg}
Consider two actors in the environment, close together.
For safety it is sufficient to predict that an obstacle exists within a boundary that encompasses both actors.
However, for this prediction (red), for each actor (green), $IoU < 0.5$ \textit{i.e.} this detection would be considered a False Positive if match were to be determined by $IoU$.
Hence $IoG$ is a better method of determining true positives for this use case.
}
\end{figure}

\textbf{{True Positives:}}
Standard object detection evaluations use an Intersection over Union ($IoU$) threshold to match predictions to groundtruth bounding boxes. For $R^3$ we use a threshold on the percentage of groundtruth bounding boxes covered by the predicted object bounding boxes (BB) \textit{i.e.} Intersection over Groundtruth ($IoG$). Figure~\ref{fig:uvsg} compares $IoU$ to $IoG$.
\begin{equation}
    IoG = \frac{\text{area(groundtruth BB)} \cap \text{area(predicted BB)} }{\text{area (groundtruth BB)}}
\end{equation}

\textbf{{Truncation and Occlusion:}}
Further from Figure~\ref{fig:uvsg}, distinguishing between different objects and detecting heavily occluded objects does not give an advantage for safety considerations, as long as the predicted bounding boxes cover the objects that exist. This is implicitly addressed by using $IoG$ to determine true positives.

\textbf{Object Classes:}
Detection of object class is not minimally required for collision avoidance as long as the object's existence is detected and worst case for its potential for motion is considered while calculating collision risk.
Collisions with any object in the AV's path, even of unknown classification, should be avoided.
Hence $R^3$ metrics do not consider object class when determining True Positives.
We do not wish to convey that classification does not have any significance. Object classes are important to determine for many path planning tasks in an AV, however purely for obstacle detection, classification is not a requirement.
However, ground truth labels can still be filtered to limit the analysis to certain classes only.
In Section~\ref{sec:usage} Pedestrians, Cycles, Vehicles and Road Signs are considered.

\textbf{Type of Object Detection System:}
In this work we heavily focus on using \rrr for the evaluation of vision-based object detection DNN but the metrics can be utilized to evaluate any system which ultimately detects the existence and location of objects around a vehicle.

The design choices above stem from the focus on collision avoidance.
\rrr cannot be used in isolation to fully compare object detection systems.
For example, Precision (Eq.~\ref{eq:pr}) is an important metric for the performance of the object detection system.
An object detection system with high \rrr but low precision may see obstacles where none exist, this could severely limit the driving performance of the vehicle even making it impossible to drive, however, it is still safe.
\rrr does not aim to replace the existing metrics rather adds additional information emphasising safety.
\section{Risk Ranking}
\label{sec:coll_risk}
\label{sec:rrr}

The category of collision risk is determined  by using ego vehicle and object state. The metrics are meant to be used with autonomous driving datasets.
Each input frame of the dataset is analyzed independently with the time instance of the current frame as $t=0$.
$x_{e}(t), x_{o}(t)$ are the position vectors of the center of the ego vehicle and object at a time $t$ respectively.
Similarly, $v_{e}(t)$ and $v_{o}(t)$ are the velocity vectors and $\theta_{e}(t)$ and $\theta_{o}(t)$ are the headings at time $t$.
The maximum possible acceleration or deceleration magnitude $ a^{max}=7.5 \: m/s^2$ is used in rest of this work, as prescribed by prior works in literature~\cite{wu2008smart, saptoadi2017suitable}.

$l_{comp}$ is the compute latency from the sensor input to the actuation command.
The latency is a property of the autonomous driving hardware and software.
Using the sample rate of the Waymo Open Dataset we use $l_{comp} = 100 ms$ in Section~\ref{sec:usage}.
$TTS$ is the time to stop for the ego vehicle from the frame input to a complete stop assuming the actuation decision was made to emergency brake. The time interval for the potential collision analysis is hence [$0$, $TTS$]
repeated in discrete time step increments $\Delta t$.
A small $\Delta t$ value approximates continuous analysis.
    
    \begin{equation}
        TTS = \frac{|v_e(0) + a^{max} * l_{comp}|}{a^{max}} + l_{comp}
        \label{eq:tts}
    \end{equation}

$d_{crit}$ is the maximum distance between the centers of the ego vehicle and the object, such that their surfaces touch at at least one point without causing any deformation of the surface.
All possible rotations of the vehicles are considered in determining $d_{crit}$, adhering to the principle of considering worst case possibilities.

\begin{figure}[tp]
\centering
\includegraphics[width=.8\linewidth,keepaspectratio]{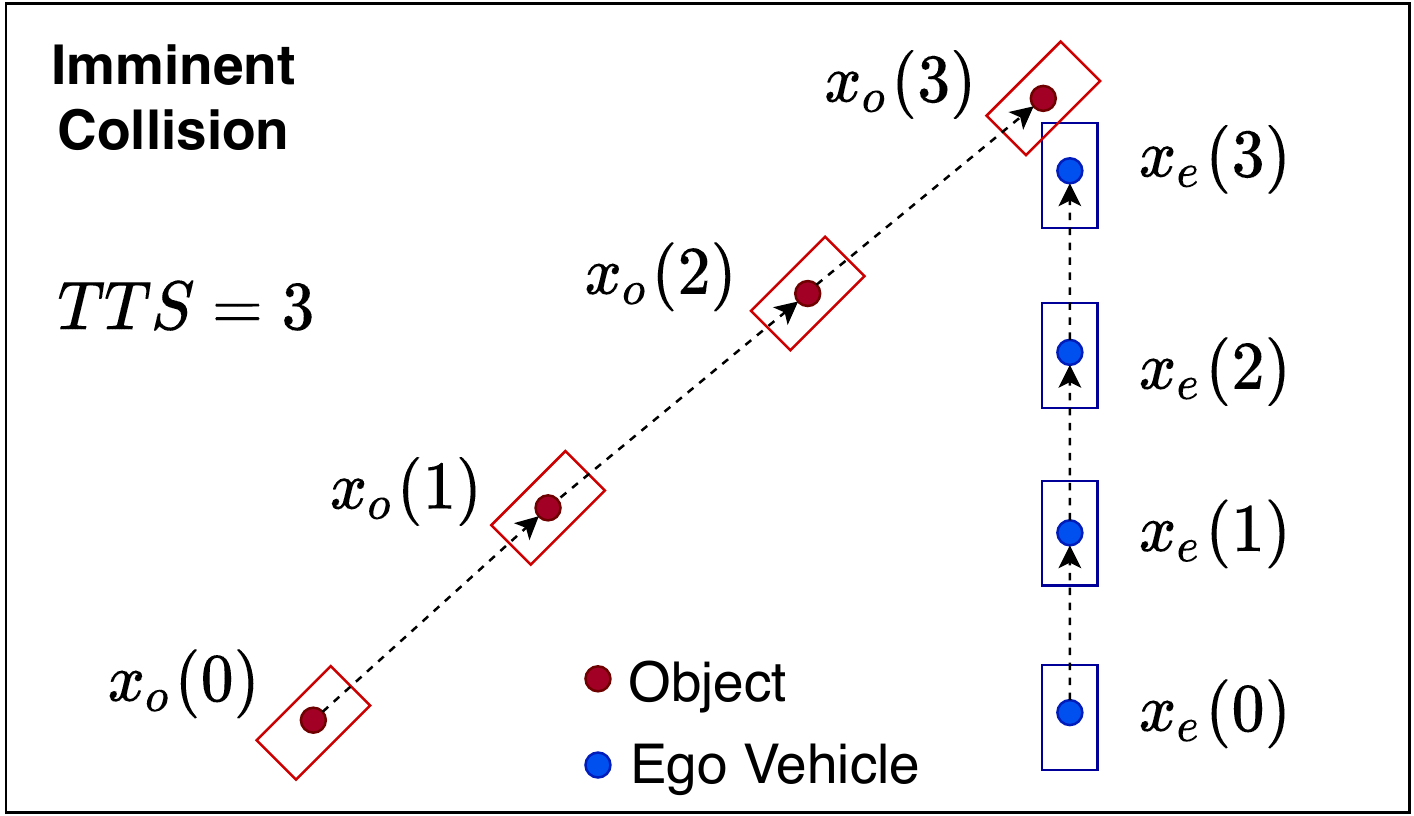}
\caption{\label{fig:cat_i}
To determine the risk of imminent collision, the trajectories of ego vehicle and object are charted over the time horizon assuming constant velocity vectors and heading. At each time instance it is checked whether the two can collide.
}
\end{figure}

\subsection{Imminent Collision}
\label{sec:imminent_collision}

To determine risk of imminent collision, \textit{i.e.} a collision that will happen if no action is taken by the ego vehicle or the object, the trajectories of both are generated within the time horizon.
$v_{e}(t)$, $v_{o}(t)$, $\theta_{e}(t)$ and $\theta_{o}(t)$ are considered constant and initial value at $t=0$ are taken from the dataset labels. If at any time, in $\Delta t$ increments, the bounding box representing ego vehicle and the object overlap, the object is ranked to pose a risk of \textit{imminent} collision.
This process is represented in Figure~\ref{fig:cat_i} with $TTS = 3 \: units$ and $\Delta t = 1 \: unit$.

\subsection{Potential Collision}
\label{sec:potential_collision}

\begin{figure}[tp]
\centering
\includegraphics[width=.95\linewidth,keepaspectratio]{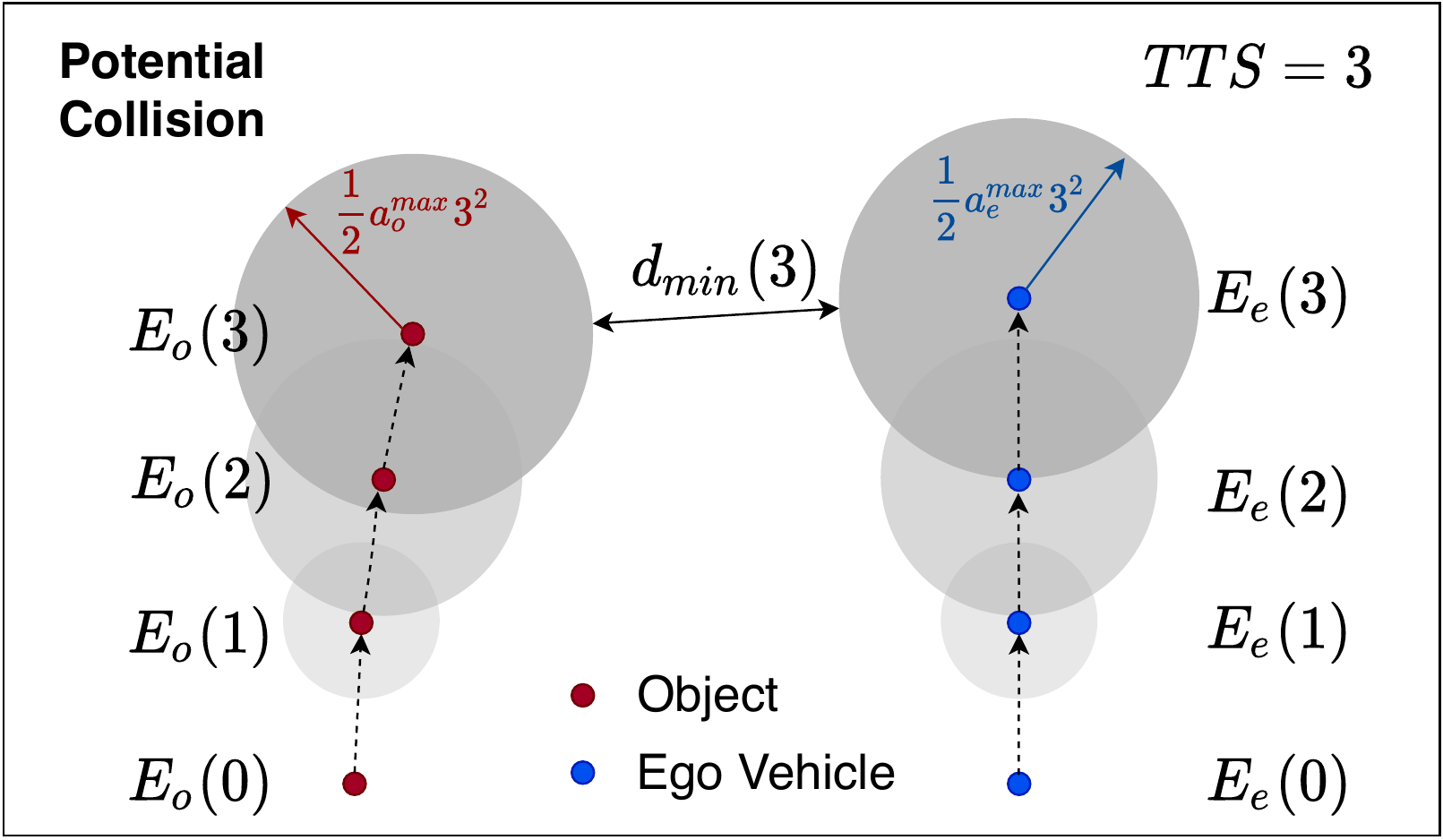}
\caption{\label{fig:cat_p}
To determine potential for collision, the existence regions of center of both the ego vehicle and object are determined over the time horizon, assuming max accelerations. The object is considered to pose a potential for collision if at any time instance the existence regions overlap or minimum distance between the existence regions $d_{min}(t) < d_{crit}$. 
}
\end{figure}

Schmidt \textit{et al.}~\cite{kamm} introduced the concept of existence region $E(t)$ of an object at a particular time $t$, defined as the set of all possible positions of the center point at a time $t$ and extended this existence region into the future via a propagation method.
Given initial $x(0)$ and $v(0)$, existence regions at time $t$ is simply a circle of radius $\frac{1}{2} a^{max} t^2$ centered at $x(0) + v(0)t$.
We derive the existence regions for both the ego vehicle and objects within $t \in [ 0, \: tts ]$.
The minimum distance between $E_e(t)$ and $E_o(t)$ at time $t$ is denoted as $d_{min}(t)$.
If at any timestep $d_{min}(t) < d_{crit}$, the object is considered to pose a risk of \textit{potential} collision as shown in Figure~\ref{fig:cat_p}.

\section{Usage}
\label{sec:usage}

\begin{figure*}[htbp]
  \includegraphics[width=\linewidth]{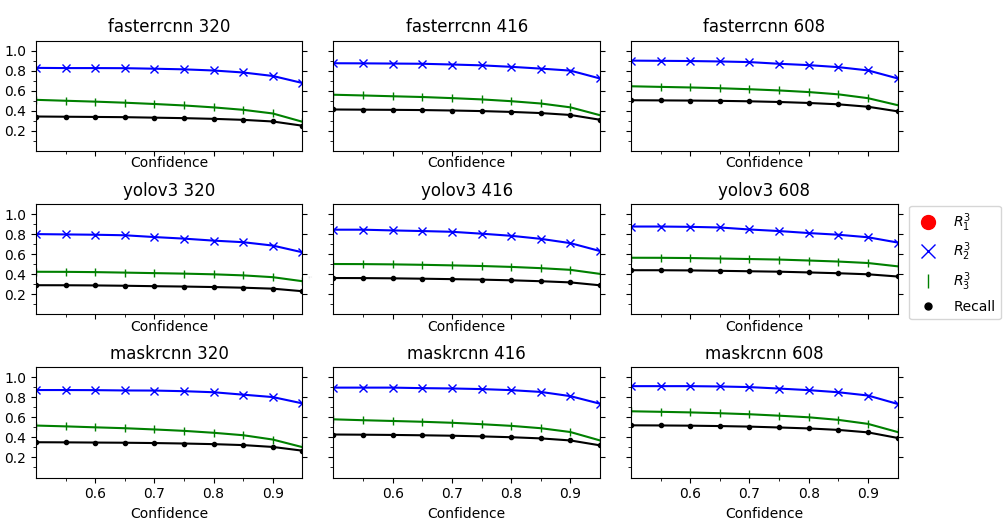}
  \caption{\label{fig:waymo_curves}
  \rrr and Recall vs Confidence for 2D object detection using Waymo Open Dataset.
  X-axis shows prediction confidence ratios $\in [0.5,.95]$ in 0.05 intervals.
  Y-axis shows the \rrr or Recall calculated using Eq.~\ref{eq:recall} for all predictions that have a confidence $\geq$ X-axis value.
  Each subplot shows the curves for a specific network $\in$ [YOLOv3, Faster R-CNN, Mask R-CNN] and input images as squares with edge size $\in$ [320, 416, 608] pixels.
  $R^3_1$ is not plotted as no objects meet this rank's criteria. \rrr uses IoG $\geq$ 0.8 to determine True Positives while Recall uses IoU $\geq$ 0.8.
  }
\end{figure*}

This section shows sample usage of \rrr metrics using three pre-trained 2D object detection networks: YOLOv3\cite{yolov3,git_yolotorch}, Faster R-CNN and Mask R-CNN~\cite{torchvision}.
Front camera images and labels are taken from 24 sequences of Waymo Open Dataset~\cite{waymo_open_dataset}.
Of the total \textbf{27973} labeled objects in these sequences, \textbf{952} are identified to pose a \textit{Potential} risk of collision.
However no objects are identified to pose a risk of an \textit{Imminent} collision.
This is an expected limitation of real world datasets and motivates the need for synthetic datasets that can  include imminently unsafe situations~\cite{kim2019crash,patel2020simulation}.
Fig.~\ref{fig:waymo_curves} shows the \rrr and Recall values for different networks and input image resolutions (320 pixels, 416 pixels, 608 pixels). 

\textbf{Discussion:}
Collision risk and detection difficulty correlate with distance between the ego vehicle and the object.
Farther objects pose lower risk of collision ($\S$\ref{sec:rrr}).
Similar correlation was observed in importance annotations by human drivers~\cite{ohn2017all}.
Farther objects are smaller and hence more difficult to detect in images, as also assumed in KITTI 2D object detection difficulty categories.
Hence in Figure~\ref{fig:waymo_curves} we see that \rrr metrics and recall show different offsets but the same trend through varying confidence limits.
Despite this apparent limitation, \rrr metrics emphasise safety critical False Negatives.

From Fig.~\ref{fig:waymo_curves} we also note that for varying resolutions of input images, Recall values see a larger difference than $R^3_2$. Smaller input sizes also come with lower computational load and latency \textit{e.g.} YOLOv3-320 has only 28\% flops and 44\% latency of YOLOv3-608 as demonstrated by Redmon and Farhadi~\cite{yolov3}.
This motivates the addition of low input resolution DNN to AV safety critical perception pipeline similar to mixed frequency networks proposed by Tesla~Inc.~\cite{patent:20200175401}.
\section{conclusion and future work}
\label{sec:future}
\label{sec:conclusion}

In this work \rrr metrics for object detection in autonomous vehicles are presented.
The usage examples for \rrr show the difference in information provided by the new metrics.
Further work is required to establish the efficacy of the metric in influencing design decisions for perception systems for AV including sensor fusion and tracking.
Another direction requiring further exploration is utilizing the metrics for redesigning the loss functions for training object detection DNN.
This is non trivial as the loss function must also consider precision and recall.
Finally, synthetic datasets that incorporate unsafe situations need to be incorporated.
While this paper presents the first step \textit{i.e.} the metric formulation, with the above addressed the metrics can truly serve object detection researchers by providing valuable information about the performance of their solutions in safety critical applications.
\section*{Acknowledgment}

The material presented in this paper is based upon work supported by
the Office of Naval Research (ONR) under grant number N00014-17-1-2783 and by
the National Science Foundation (NSF) under grant numbers CNS 1646383, CNS 1932529 and CNS 1815891.
Marco Caccamo was also supported by an Alexander von Humboldt Professorship
endowed by the German Federal Ministry of Education and Research. Any
opinions, findings, and conclusions or recommendations expressed in
this publication are those of the authors and do not necessarily
reflect the views of the sponsors.

\bibliography{ref}
\end{document}